\documentclass[conference]{IEEEtran}
\usepackage{times}

\usepackage{authblk}
\usepackage{comment}
\usepackage[numbers]{natbib}
\usepackage{multicol}
\usepackage[bookmarks=true]{hyperref}
\usepackage{amsmath}
\usepackage{amssymb}
\usepackage{graphicx}
\usepackage{tabu}
\usepackage{adjustbox,multirow}
\usepackage{makecell}
\usepackage{caption}
\usepackage{subcaption}
\usepackage[ruled,vlined]{algorithm2e}
\SetKwInput{KwInput}{Input}                
\SetKwInput{KwOutput}{Output}              
\usepackage{algorithmicx,algpseudocode,float}
\usepackage{pgfplots}
\pgfplotsset{compat=1.9}
\usepackage{bm}

  {
      \newtheorem{assumption}{Assumption}
      
  }

\usepackage{longtable}
\usepackage{makecell, cellspace, caption}
\usepackage{array}
\newcolumntype{L}[1]{>{\raggedright\let\newline\\\arraybackslash\hspace{0pt}}m{#1}}
\newcolumntype{C}[1]{>{\centering\let\newline\\\arraybackslash\hspace{0pt}}m{#1}}
\newcolumntype{R}[1]{>{\raggedleft\let\newline\\\arraybackslash\hspace{0pt}}m{#1}}

\begin{document}

\title{Learning Geometric Constraints \\
in Task and Motion Planning}

\author[1]{Tianyu Ren}
\author[2]{Alexander Imani Cowen-Rivers}
\author[2]{Haitham Bou Ammar}
\author[1]{Jan Peters}
\affil[1]{Computer Science Department, Technische Universität Darmstadt}
\affil[2]{Huawei R\&D London}

\maketitle

\begin{abstract} 
Searching for bindings of geometric parameters in task and motion planning (TAMP) is a finite-horizon stochastic planning problem with high-dimensional decision spaces.A robot manipulator can only move in a subspace of its whole range that is subjected to many geometric constraints. A TAMP solver usually takes many explorations before finding a feasible binding set for each task. 
It is favorable to learn those constraints once and then transfer them over different tasks within the same work space. We address this problem by representing constraint knowledge with transferable primitives and using
Bayesian optimization (BO) based on these primitives to guide binding search in further tasks. Via semantic and geometric backtracking in TAMP, we construct \textit{constraint primitives} to encode the geometric constraints respectively in a reusable form. Then we devise a BO approach to efficiently utilize the accumulated constraints for guiding node expansion of a MCTS-based binding planner. We further compose a transfer mechanism to enable free knowledge flow between TAMP tasks. Results indicate that our approach reduces the expensive exploration calls in binding search by 43.60to 71.69 when compared to the baseline unguided planner.
\end{abstract}

\section{Introduction}
\textit{Task and Motion Planning (TAMP)} is a framework aiming to enable robotic reasoning and acting in multi-stage long-horizon manipulation and mobility tasks.
TAMP solvers typically adhere to a two-level execution process. First, high-level planners establish a sequence of symbolic operators (skeletons) that allow arriving at a (symbolic) goal state. With those sequences at hand, low-level motion planners then bind symbolic operators to metric motion parameters (bindings), effectively rendering geometrically feasible strategies for the environment.

Although successful in many instances~\cite{kaelbling2011hierarchical,srivastava2014combined,dantam2016incremental,garrett2020pddlstream,garrett2020integrated}, symbolic binding in TAMP is challenging, requiring resolutions to large-scale constraint stochastic sequential decision-making problems. To illustrate, consider the grasping example in Figure~\ref{fig:tree} in which a robot needs to transport a \texttt{body} from \texttt{region1} to \texttt{region2} while avoiding environmental obstacles, e.g., \texttt{obstacle1} or \texttt{obstacle2}. The feasible skeleton of Table~\ref{tab:skeleton} requires binding decisions that, first, choose a grasping direction $\texttt{\#dir1} \in \{\texttt{front},\texttt{right},\texttt{top},\texttt{back},\texttt{left}\}$ and then opt a target pose \texttt{\#pose12} $\in [0,1]^3$. Those bindings are sequential such that a choice of $\texttt{\#dir1}$ alters allowable pose states that can subsequently affect later decision stages. Existing approaches trend to address such binding search by modelling each task as a black-box optimization problem~\cite{kim2020monte,yee2016monte,kim2019learning}. In this case, different environmental constraints (e.g., those induced by \texttt{obstacle1} and \texttt{obstacle2}) are merged together into a single objective that is specified for a task instance. Therefore, experience about these constraints cannot be shared by a second task.


To promote knowledge sharing in binding search, in this paper, we represent environmental constraints respectively by \textit{constraint primitives (CPs)}. Based on constraint primitives, we devise a Bayesian Optimization algorithm to effectively guide the exploration binding search in TAMP.
The main contribution of this work is threefold:
\begin{itemize}
    \item we create constraint primitives each of which represents a piece of objective geometric knowledge of the environment via semantic and geometric backtracking;
    \item we propose an efficient BO algorithm based on constraint primitives to guide the exploration in the decision tree of binding search;
    \item we devise a transfer learning mechanism to generalize accumulated constraint primitives to new tasks with zero effort.
\end{itemize}

In three robot manipulation tasks, we show that the proposed method can improve binding search efficiency over planners with quasi-random samplers. We also provide a comparison with another Bayesian optimization approach that does not use the constraint primitive representation to show the significant improvement with transferred knowledge.

\section{Related work}
Most existing studies solve binding search by optimizing a single objective function with respect to a specific task. Geometric constraints of the environment are implicitly modelled together in the objective. For addressing such a finite-horizon optimization problem.
MCTS (Monte Carlo tree search) or UCT (Upper Confidence bounds for Trees) \cite{kocsis2006improved} are common choices. 
In \cite{kim2020monte}, VOOT (Voronoi optimistic optimization applied to trees) performs value-driven sampling of the continuous binding space. It requires a deterministic objective
function so it is infeasible for most TAMP systems where sample-based motion generators (e.g., RRT \cite{lavalle1998rapidly}) are extensively used.
Kernel Regression UCT proposed in \cite{yee2016monte} enables information sharing between similar binding decisions through kernel regression, but it does not provide any convergence or completeness guarantees.
In eTAMP \cite{ren2021extended}, PW-UCT (Upper Confidence bounds for Trees with Progressive Widening) \cite{couetoux2011continuous,auger2013continuous} is used to address the stochastic transition dynamics in planning. It ensures probabilistic completeness in binding searching by observing the PW laws prescribed in \cite{auger2013continuous}. This approach however relies on random samplers for node expansion in UCT. The resultant exploration is uninformative and inefficient. In addition, as with other MCTS methods, PW-UCT have difficulties transfer its experience to other tasks. It has to start from scratch for each TAMP task even though there is apparent overlap in geometric constraints (e.g., $T_\text{desk}$ and $T_\text{deskP}$ in Fig.\,\ref{fig:transDesk} share the same constraint with \texttt{obstacle3}). 
In \cite{kim2019learning}, the authors formulate binding search as a black-box function optimization problem and propose a experience-based UCB (Upper Confidence Bound) algorithm, BOX, to guide binding search. 
By maintaining a score matrix with columns divided by binding decisions and rows divided by task instances, BOX can reason with the correlation information between different task instances so that it gets some capability to accumulate binding experience and generalize it to unseen instances. However, this framework only works with discrete binding spaces, and it only enables knowledge sharing between different instances (with the same skeleton) instead of different tasks (e.g., $T_\text{desk}$, $T_\text{deskP}$, $T_\text{regrasp}$ in Fig.\,\ref{fig:transDesk}). Moreover, the construction of the score matrix requires substantial engineering and training before it can be used for each task. 

Instead of merging into a single objective, environmental constraints can be respectively modeled and satisfied during binding search. With hand-coded constraints, \cite{lozano2014constraint} formulates a pick-and-place task as a constraint satisfaction problem (CSP) with discrete binding spaces and solves it with off-the-shelf CSP solvers. By dependency analysis in semantic and geometric spaces, \cite{lagriffoul2016combining} proposes a culprit detection mechanism to automatically identify constraints in the robot environment. To make constraints more general and reusable, authors in \cite{lagriffoul2016combining} use typed symbols instead of concrete instances to describe constraints.
Unfortunately, this method also requires discretization of the binding space, which adds difficulties in application to practical robot tasks. 
We are interested in knowledge transfer mechanism for binding search that supports decision making in both continuous and discrete parameter space. It should have no negative effects on the planner completeness and should require no special training phases.

\section{Background}

\begin{table*}[t!]
    \caption{\small A feasible skeleton for $T_\text{desk}$ with its operators grouped by layers of a decision tree.}     \label{tab:skeleton}
\centering
\begin{tabular}{c|c}
    \hline
    \makecell{Tree node layer} & \makecell{Skeleton operators}\\
    \hline
    \texttt{decision1} & \makecell[l]{$\texttt{Sample-grasp(body)} \to \texttt{\#dir1}$}\\
    \hline
    \texttt{transition1} & \makecell[l]{
    $\texttt{Inv-kin(body,pose0,\#dir1)} \to \texttt{\#config1}$ \\
    $\texttt{Plan-motion(config0,\#config1)} \to \texttt{\#traj01}$ \\
    $\texttt{Move-Pick(body,\#traj01)}$}
    \\
    \hline
    \texttt{decision2} & \makecell[l]{$\texttt{Sample-pose(body,region2)} \to \texttt{\#pose12}$} \\
    \hline
    \texttt{transition2} & \makecell[l]{
    $\texttt{Inv-kin(body,\#pose12,\#dir1)} \to \texttt{\#config2}$ \\
    $\texttt{Plan-motion(\#config1,\#config2)} \to \texttt{\#traj12}$ \\
    $\texttt{Move-Place(body,\#traj12)}$} 
    \\
    \hline
    \end{tabular}
    \vspace{-0.5cm}
\end{table*}

As the starting point of binding search, we assume a feasible skeleton for $T_\text{desk}$ as shown in Table\,\ref{tab:skeleton}, where \texttt{pose0} is the initial pose of \texttt{body} on \texttt{region1}, and \texttt{config0} is the initial configuration of \texttt{robot}. 
The motion parameters marked by \texttt{\#} are those open decisions that demand concrete \textit{bindings}. 

\subsection{Zero-shot Task Generalization}

The goal of zero-shot task generalization is to achieve task goals that are not seen during training [24–
26].For evaluation we consider a zero-shot generalization
setup [25, 26] where only a subset of the task goals is available during training, and the agent has to achieve a disjoint set of test task goals.

\subsection{UCT-based Binding Search}\label{sub:uct}

\begin{figure}
    \centering
    
        \begin{subfigure}[b]{0.22\textwidth}
        \centering
        \includegraphics[width=1\textwidth]{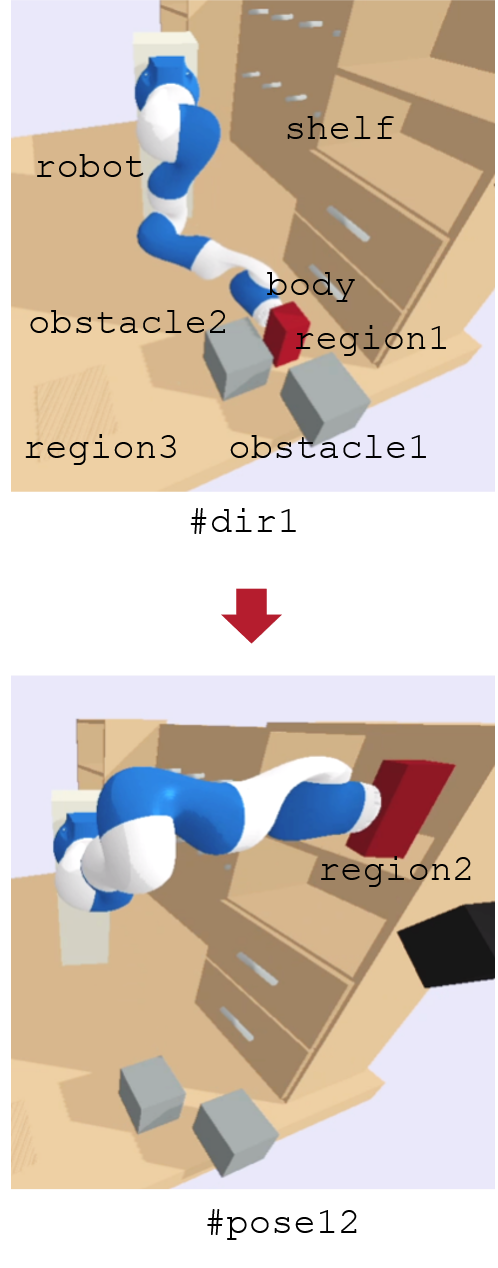}
            \vspace{-0.18cm}
        \caption{$T_\text{desk}$}
        \label{fig:task_desk}
    \end{subfigure}
    \hfill
    \begin{subfigure}[b]{0.5\textwidth}
        \centering
        \includegraphics[width=1\textwidth]{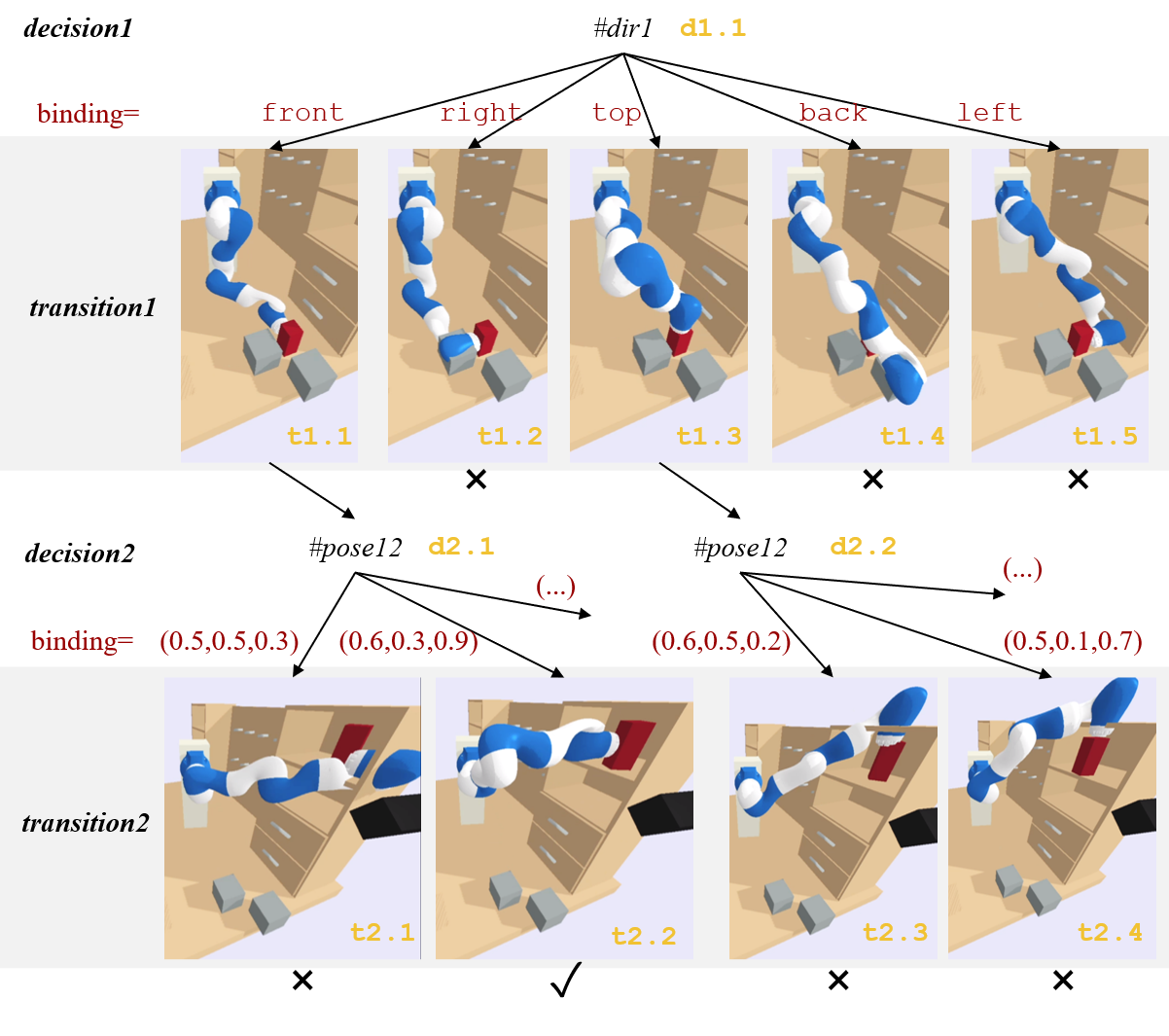}
            \vspace{-0.18cm}
        \caption{The decision tree for $\text{BINDING}(T_\text{desk},\langle \texttt{\#dir1,\#pose12} \rangle)$}
        \label{fig:decision_tree}
    \end{subfigure}
    
    \caption{\small $T_\text{desk}$ with motion parameters $\langle \texttt{\#dir1,\#pose12} \rangle$ to be bound by the decision tree.}
    \label{fig:tree}
    \vspace{-0.5cm}
\end{figure}

Following \cite{auger2013continuous}, we model the binding search process as a decision tree. In a similar way of how \textit{solution constraints} are defined in \cite{kim2019learning}, we group the skeleton operators into two alternative layers (see Table\,\ref{tab:skeleton}): the \textit{transition layer} and the \textit{decision layer}. Operators in decision layers, such as \texttt{Sample-grasp} and \texttt{Sample-pose}, always generate decisions with long-lasting effects and they are especially crucial to the solution.
We note the binding search problem as $\text{BINDING}(Task,\langle \texttt{\#d1,\#d2,...} \rangle)$. An decision tree for $\text{BINDING}(T_\text{desk}, \langle \texttt{\#dir1,\#pose12} \rangle)$ can be plotted as Fig.\,\ref{fig:tree}.  

The PW-UCT algorithm for binding search is outlined in Alg.\,\ref{alg:uct} (please see details in App.\,\ref{app:a}).
The planner should search for feasible bindings $\bm{x}^{\star}$ (e.g., $\bm{x}^{\star}=\langle \texttt{front,(0.6,0.3,0.9)} \rangle$ in Fig.\,\ref{fig:tree}) until the maximum attempts $N_{rollout}$ is reached. Before $\bm{x}^{\star}$, each UCT rollout is concluded by a terminal state. A termination means at least one geometric constraint is violated, and binding search must restart from the root. A tree node at termination state will receive a reward $r = 0.1\left(d_\text{end}/d_\text{total}\right)+r_\text{end}$ \label{eq:reward}, where $d_\text{end}$ is the termination depth in UCT $d_\text{total}$ is the total tree depth. $r_\text{end}=1$ when a feasible binding list is found, otherwise $r_\text{end}=0$. This reward encourages the planner to go deeper in the decision tree and to find more feasible bindings.

On SAMPLE-NEW-CHILD$(node)$, a new child node will be expanded from $node$ with a binding decision generated by a random sampler. 
Otherwise an existing child node of $node$ will be selected by UCB criterion. The balance between sampling new children and selecting existing children is regulated by EXPANDABLE-BY-PW$(node)$. The basic PW idea is to limit the number of visits for existing nodes. Despite its probabilistic completeness in theory, we have observed in practice slow convergence to feasible solutions with this approach. It is mainly due to the unguided random samplers used for exploration. Without any information about the geometric constraints of the environment, these samplers just enumerate bindings by brute force. We propose to address this problem by replacing random samplers in Alg.\,\ref{alg:uct} with Bayesian optimisation routines.

\begin{algorithm}
\DontPrintSemicolon
\KwInput{$N_{rollout}$}
$node \leftarrow root$\\
\While{$count<N_{rollout}$}
{
    \While{$node$ is not terminated and $node$ is not su}
    {
        ADD-VISIT($node$)\\
        \eIf{EXPANDABLE-BY-PW$(node)$}
        {
            $node \leftarrow \text{SAMPLE-NEW-CHILD}(node)$ 
        }
        {
            $node \leftarrow \text{SELECT-CHILD-BY-UCB}(node)$
        }
    }
    BACK-PROPAGATE$(node)$\\
    \vspace{1.0mm}
    $node \leftarrow root$\\
    $count \leftarrow count+1$\\
    $\bm{x}^{\star}=$GET-FEASIBLE-BINDINGS$(root)$ \\
    \If{$\bm{x}^{\star}$ is not None}
        {
            \textbf{return} $\bm{x}^{\star}$
        }
}
\textbf{return} $None$
\caption{\small UCT Binding Search}\label{alg:uct}
\end{algorithm}



Commonly-used random samplers in tree node expansion (SAMPLE-NEW-CHILD in Alg.\,\ref{alg:uct}) lead to unguided explorations in UCT \cite{couetoux2011continuous,auger2013continuous}. Thus the binding planner normally requires a large amount of simulation rollouts before a feasible solution $\bm{x}^{\star}$ is found. In comparison, the framework we purpose CP-BO exploits each failure point during the tree search and accumulate it as a data point. These points are stored in a global dataset. 

\subsection{Bayesian Optimization}
As noted earlier, TAMP requires expensive exploration step in UCT. In the context of TAMP we want to explore multiple constraints till we are feasible. This optimisation problem differs from typical BO as we have fine-grained information as to whether each individual constraint primitives in the simulator was satisfied. Thus, for $n$ constraint primitives we have both the overall information of whether the bindings were feasible for the task $y=f(\bm{x})$ and for each primitive constraint $i$ we receive additional information ($y^\texttt{CP}_i$) as to whether the constraint was satisfied. Note, in order for feasibility within a task, all constraints that compromise this task must be satisfied. Because of this additional information, we can frame a novel Bayesian Optimisation problem as shown in Eq.~\ref{Eq:BB}.

\begin{equation}
\label{Eq:BB}
 \arg\max_{\bm{x} \in \mathcal{X}} f(\bm{x}),
\end{equation}
with $\bm{x}$ denoting a binding list of motion parameters, $\mathcal{X}$ a bounded binding domain and $f: \mathcal{X} \rightarrow \mathcal{R}$ determines whether a feasible binding is found ($\mathcal{P}(X=\text{Feasible})=1.0$) for a skeleton with respect to all the geometric constraints. 


To achieve this goal, BO algorithms operate in two steps. In the first, a Bayesian surrogate model is learned, while in the second an acquisition function is maximised to determine new bindings. 


\textbf{Step 1: }To learn a Bayesian regression surrogate model, one typically places a GP prior on the latent function, $f(\cdot)$, which is fully specified through a mean function, $m(\bm{x})$, and a covariance function or kernel $k_{\bm{\theta}}(\bm{x}, \bm{x}^{\prime})$ with $\bm{\theta}$ representing kernel hyper-parameters. The model specification is completed by defining a likelihood. Here, practitioners typically assume that observations $y_{l}$ adhere to a Gaussian noise model such that $y_l = f(\bm{x}_l) + \epsilon_l$ where $\epsilon_l \sim \mathcal{N}(0, \sigma_{\text{noise}}^{2})$. This, in turn, generates a Gaussian likelihood of the form $y_l | \bm{x}_{l} \sim \mathcal{N}(f_l, \sigma_{\text{noise}}^{2})$ where we use $f_l$ to denote $f(\bm{x}_{l})$ with $f(\bm{x}) \sim \mathcal{G}\mathcal{P}(m(\bm{x}), k_{\bm{\theta}}(\bm{x}, \bm{x}^{\prime}))$. To learn a Bayesian classification model, one must additionally warp the the output to be between [0,1] via the standard Normal CDF $\Phi(x)$, using a variational GP as we no longer preserve an analytic form our exact GP objective.

\textbf{Step 2: }To determine bindings $\bm{x}^{\star}$, we typically maximise an acquisition function $\alpha$, such as Upper Confidence Bound (UCB) using either an evolutionary search (ES) or gradient approach~\cite{Wilson2018a}. However, for our specific setting of multiple constraints which involve both continuous and binary constraints, they do not work out the box, thus we introduce a new acquisition function in Sec.~\ref{sec:methodsb}.


\begin{equation}
    \bm{x}^{\star} = \arg \max_{\bm{x}} \alpha^{\bm{\theta}}_{\text{UCB}}(\bm{x}|\mathcal{D}), \ \ \ \alpha_{\text{UCB}}^{\bm{\theta}}(\bm{x}|\mathcal{D}) = \mu(\bm{x}; \boldsymbol{\theta}) + \sqrt{\beta} \sigma(\bm{x};\boldsymbol{\theta})
\end{equation}\label{eq:ucb}


Where $\beta$ is a hyper-parameter that controls the exploration–exploitation trade-off. In this work we use the ES method NSGA-II, to maximise the acquisition functions. 

\section{Representing \& Utilising Constraint Primitive Knowledge}\label{sec:methods}

To achieve guided exploration in binding tree search, firstly the geometric constraints experienced in the previous simulation rollouts should be properly represented as transferable knowledge that is easy to store and retrieve, as described in Sec.~\ref{sec:methodsa}. In~\ref{sec:methodsb} we purpose our method Constraint Primitive Bayesian Optimisation (CP-BO) to utilise the stored information from Sec.~\ref{sec:methodsa} to reduce calls to the (expensive) simulator. Lastly, in~\ref{sec:methodsc} we then describe how to transfer the knowledge stored from Sec.~\ref{sec:methodsa} across tasks. 


\subsection{Causal Graph: Representing CP Knowledge}\label{sec:methodsa}

To transfer knowledge across tasks, we must make environmental constraints independent from specific tasks and represent them as objective knowledge.
We propose to use \textit{constraint primitives (CP)} to model the respective geometric restrictions of the environment. For example in Fig.\,\ref{fig:tree}, the planner should learn from the failure at \texttt{t1.2}: the unsuccessful inverse kinematic solver \texttt{Inv-kin} indicates that  \texttt{right} is not a good binding for \texttt{\#dir1} when \texttt{body} is at \texttt{pose0} due to the collision with \texttt{obstacle2}. We describe a CP type by four elements: the failure vertex (e.g., \texttt{Inv-kin}), the the responsible decision vertexes (e.g., \texttt{\#dir1}), the relevant context vertexes (e.g., \{\texttt{obstacle2,pose0}\}), and other connection vertexes.

\begin{figure}[t!]
    \centering
    \begin{subfigure}[b]{0.49\textwidth}
        \centering
        \includegraphics[width=0.8\textwidth]{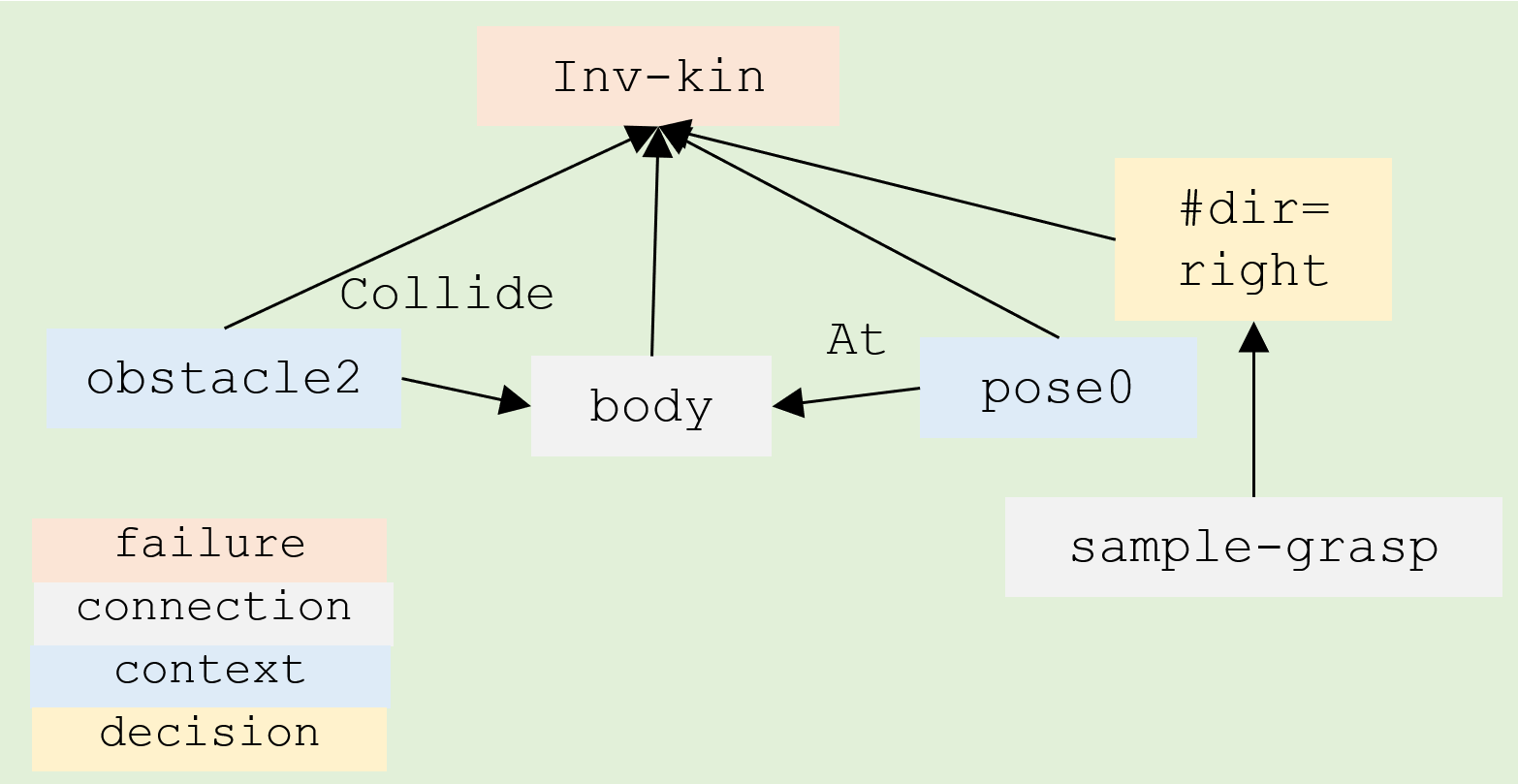}
        \caption{$\texttt{t1.2} \in \texttt{CP}_\texttt{a}(\langle \texttt{\#dir} \rangle,\{ \texttt{obstacle2,pose0}\})$}
        \label{fig:CP12}
    \end{subfigure}
    \hfill
    \begin{subfigure}[b]{0.49\textwidth}
        \centering
        \includegraphics[width=0.8\textwidth]{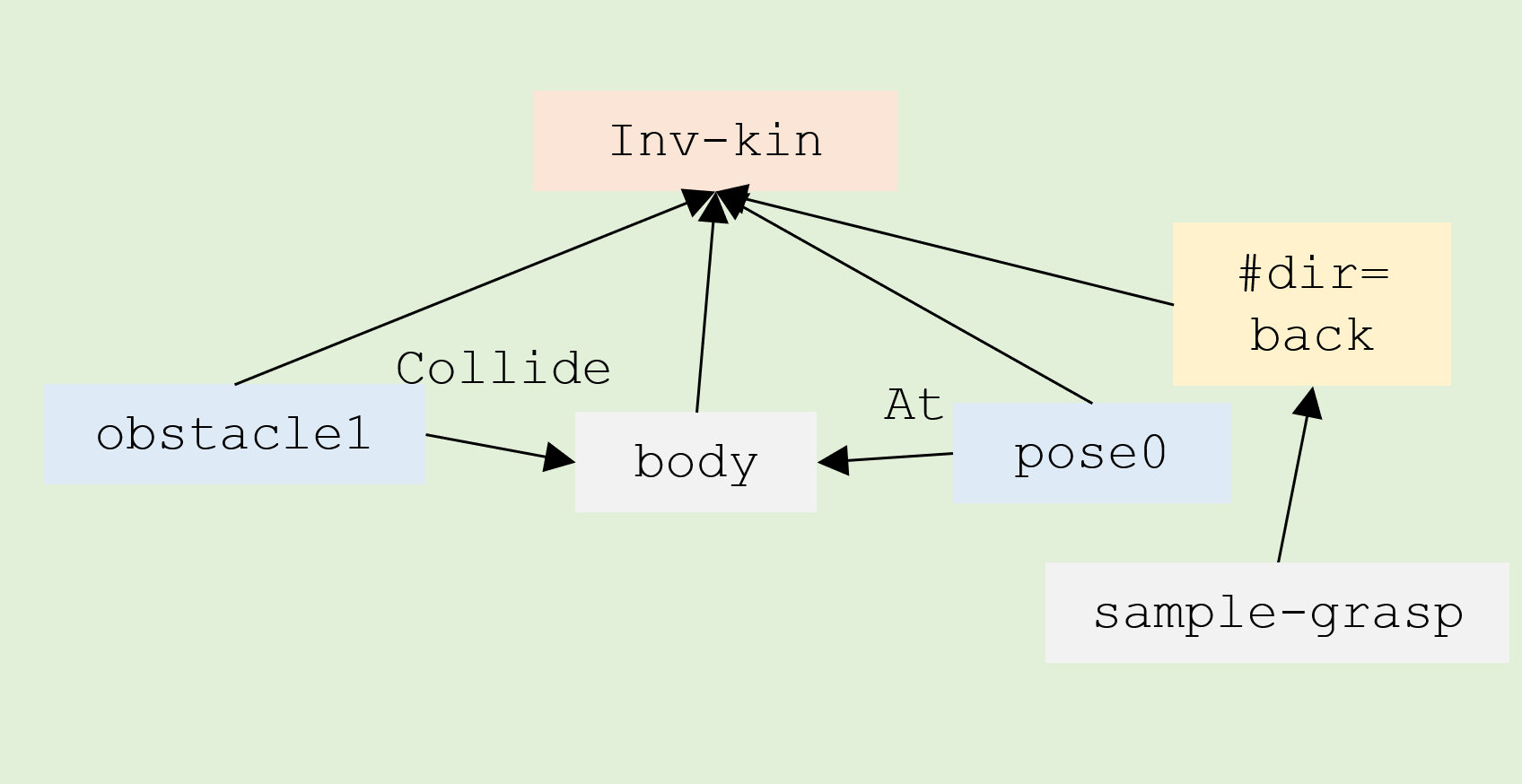}
        \caption{$\texttt{t1.4} \in \texttt{CP}_\texttt{b}(\langle \texttt{\#dir} \rangle,\{ \texttt{obstacle1,pose0}\})$}
        \label{fig:CP14}
    \end{subfigure}
    \vfill
        \begin{subfigure}[b]{0.49\textwidth}
        \centering
        \includegraphics[width=0.8\textwidth]{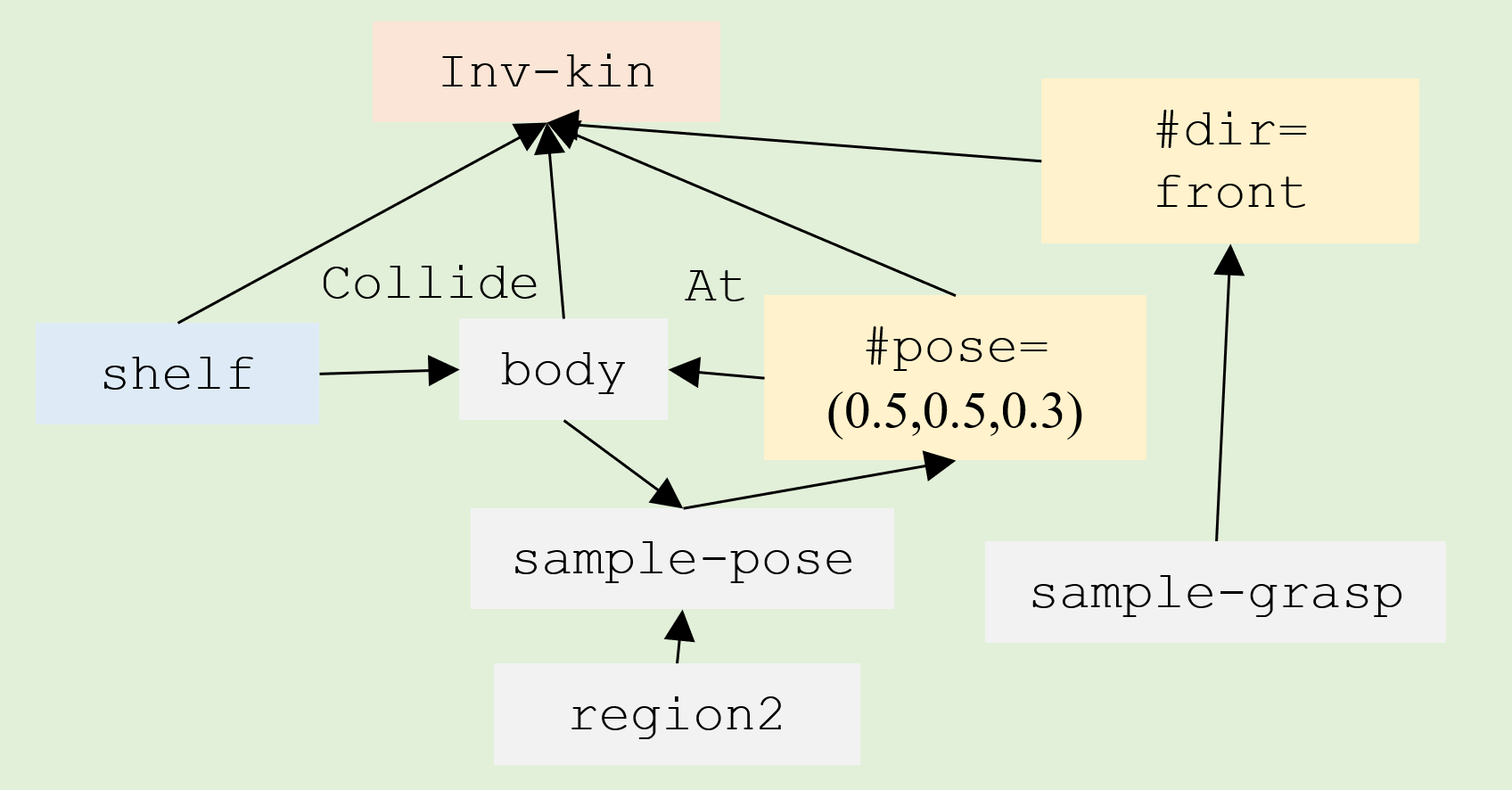}
        \caption{$\texttt{t2.1} \in \texttt{CP}_\texttt{d}(\langle \texttt{\#dir,\#pose} \rangle,\{ \texttt{shelf}\})$}
        \label{fig:CP21}
    \end{subfigure}
    \hfill
    \begin{subfigure}[b]{0.49\textwidth}
        \centering
        \includegraphics[width=0.8\textwidth]{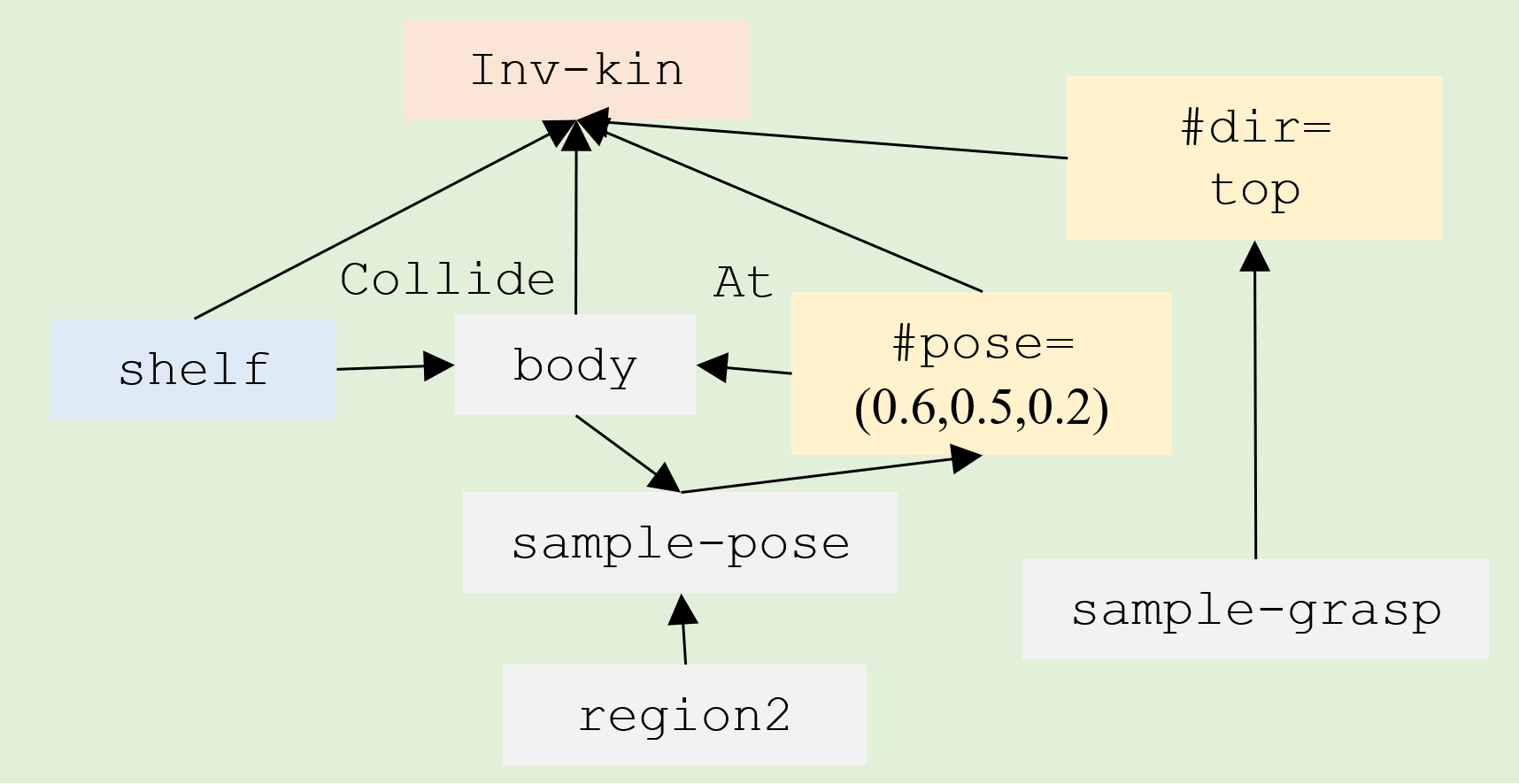}
        \caption{$\texttt{t2.3} \in \texttt{CP}_\texttt{d}(\langle \texttt{\#dir,\#pose} \rangle,\{ \texttt{shelf}\})$}
        \label{fig:CP23}
    \end{subfigure}
    \caption{\small Constraint primitive points are represented by directional graphs extracted from the semantic causal relationships provided by skeletons. $\texttt{CP}_\texttt{a}$, $\texttt{CP}_\texttt{b}$, $\texttt{CP}_\texttt{d}$ represent three different CP types.}
    \label{fig:CP_type}
            \vspace{-0.6cm}
\end{figure}

\textbf{Encode CP by Causal Graph:} A directional graph is a natural choice for representing the causal relationships between the four elements of a CP type. With directional graphs, we can compare CP types via examining the isomorphism of their graphs and their corresponding vertexes. We plot in Fig.\,\ref{fig:CP_type} the causal graphs of some failure nodes in Fig.\,\ref{fig:decision_tree}. 
Whenever a failure is detected during binding searching, we utilize the skeleton to backtrack the semantic relationships starting from the failure vertex, and meanwhile we use geometric dependency chains rendered by the simulator to backtrack  obstacles that result in collisions. The directional edges in a causal graph represent input-output as well as the collision relationships. 

\textbf{CP types and CP points: } We further hash the causal graph by Weisfeiler Lehman method \cite{shervashidze2011weisfeiler} into a hexadecimal string and use it to constitute a unique ID for a certain CP type: the type ID. CPs with the same type ID are considered belong to the same CP type.
We formally define a CP point of \texttt{typeID} as \eqref{eq:cp_point}. It indicates that the bindings of $\bm{x}=\langle binding_1,binding_2,...\rangle$ in $\{context_1,context_2,...\}$ is evaluated as $\bm{y}^{\texttt{CP}}_\texttt{typeID}$. Scalar $\bm{y}^{\texttt{CP}}_\texttt{typeID}$ measures the severity of \texttt{typeID} being breached. In its simplest implementation, we set $\bm{y}^{\texttt{CP}}_\texttt{typeID}=-1$ for feasible CP points and $\bm{y}^{\texttt{CP}}_\texttt{typeID}=0$
 for the infeasible.

\begin{equation}\label{eq:cp_point}
\begin{split}
\bm{y}=\texttt{CP}_\texttt{typeID}(\langle binding_1,binding_2,...\rangle,
    \{context_1,context_2,...\}) 
\end{split}
\end{equation}

The granularity of CP type is determined by the level of details encoded in type ID. The more semantic and geometric features are included in type ID, the finer the CP are subdivided. A definition should be detailed enough to divide the constraints into groups in each of which the data points are comparable to each other and useful covariance information can be extracted by Bayesian optimization.
For example in Fig.\,\ref{fig:CP_type}, $\texttt{CP}_\texttt{a}$ should be distinguished from $\texttt{CP}_\texttt{b}$ since it is subjected to a different obstacle; while $\texttt{CP}_\texttt{c}$ should is distinguished from $\texttt{CP}_\texttt{d}$ as it has different decision arguments. 
On the other hand, the CP representation should be general enough to enable knowledge transfer across as many as tasks. 

\begin{figure*}[t!]
    \centering
    \includegraphics[width=0.98\textwidth]{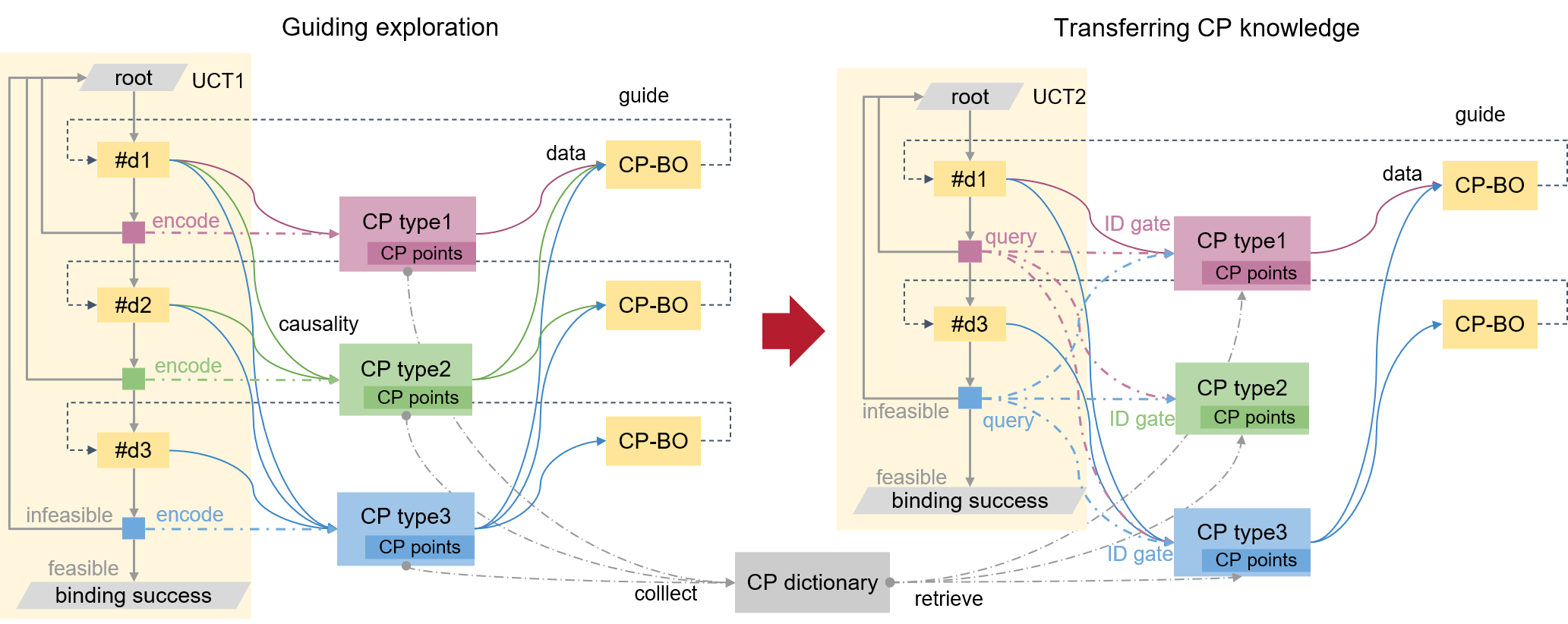}
    \caption{\small CP-BO assists binding search by guiding exploration (left) and transferring knowledge (right).}
    \label{fig:tranfer_cp}
    \vspace{-0.5cm}
\end{figure*}

For the TAMP tasks of concern, we can use a global CP dictionary $\mathcal{D}=\{\text{type ID} :=\ \mathcal{D}^{\texttt{CP}}_\text{type ID}\}$ to store CP points by CP types. With this knowledge container, each failure in binding search is expected to contribute to the total understanding of the environment.
During binding search (see Fig\,\ref{fig:tranfer_cp} (left)), a failure point $\texttt{point}_\texttt{x}$ is encoded to a type ID $\texttt{type}_\texttt{x}$. If $\texttt{type}_\texttt{x}$ already exists in $\mathcal{D}$, we add $\texttt{point}_\texttt{x}$ to the data set of $\texttt{type}_\texttt{x}$; otherwise we create $\texttt{type}_\texttt{x}$ in $\mathcal{D}$ and initialize it with $\texttt{point}_\texttt{x}$.

\subsection{Bayesian Optimisation Over Constraint Primitive Knowledge}\label{sec:methodsb}

When constraint primitives are collected, they can support binding decision making thereafter.

\textbf{Acquisition Function: } Given a set of $n$ constraints $\texttt{CP}_1, \ldots, \texttt{CP}_n$ and corresponding feasible sets of constraint values $\texttt{F}_1, \ldots, \texttt{F}_n$, and corresponding constraint datasets $\mathcal{D}^\texttt{CP}_0,\ldots, \mathcal{D}^\texttt{CP}_n$,  we notice that our main objective~\ref{Eq:BB} can be further decomposed with individual constraint primitives, specifically $\mathcal{P}^{\text{Feasible}} = \prod_i^n \mathcal{P}(\texttt{CP}_\texttt{i} \in \texttt{F}_\texttt{i} \mid \bm{x},\mathcal{D}^\texttt{CP}_i)$. Inspired by this decomposition, we introduce a novel acquisition function named Minimum Probability of Feasibility (MPF) that allows us to successfully find bindings with high probability of satisfying \textit{all} constraints. 

\begin{align*}
    \alpha^{\bm{\theta}}_{\text{MPF}}(\bm{x}|\mathcal{D}) &= \min \Bigg[\mathcal{P}(\texttt{CP}_1 \in \texttt{F}_1), \ldots, \mathcal{P}(\texttt{CP}_n \in \texttt{F}_n) \Bigg],
\end{align*}

Where we use $\mathcal{P}(\texttt{CP}_1 \in \texttt{F}_1)=\mathcal{P}(\texttt{CP}_1 \in \texttt{F}_1 \mid \bm{x},\mathcal{D})$ for compactness. To give further intuition as to the benefit of this formulation, once we maximised our acquisition and found an $\bm{x}^{\star}$, there exists a $\bm{p}^{\star}$ such that $\alpha^{\bm{\theta}}_{\text{MPF}}(\bm{x}^{\star})=\bm{p}^{\star}$, where $\mathcal{P}(\texttt{CP}_\texttt{k} \in \texttt{F}_\texttt{k}) \geq \bm{p}^{\star}$ for $k=1,\ldots\,n$. With this new acquisition function, we can now purpose our full Bayesian Optimisation algorithm CP-BO (Alg.~\ref{alg:cpbo}) to suggest bindings for UCT. Additionally, perform BO directly on the UCT reward function (detailed in~\ref{sub:uct}) using the UCB acquisition~\ref{eq:ucb}, which we name MCTS-BO.

\begin{algorithm}[h!]
\DontPrintSemicolon
\KwInput{Total number of constraints $N$, constraint dataset $\mathcal{D}^\texttt{CP}_i = \{\bm{x}, \bm{y}^\texttt{CP}_i\}_{l=1}^{n_{0}}$, $\alpha^{\bm{\theta}}_{\text{MPF}}$.}
\While{Not Feasible}
{   
    Fit a surrogate model to each constraint $i$'s dataset $\mathcal{D}^\texttt{CP}_i$. \\
    Find new bindings $\hat{\bm{x}}$ by maximising $\alpha^{\bm{\theta}}_{\text{MPF}}$. \\
    Evaluate new bindings by querying the simulator to acquire [$\hat{\bm{y}}^\texttt{CP}_0, \ldots, \hat{\bm{y}}^\texttt{CP}_n ]= f(\hat{\bm{x}})$. \\
    Update each constraint $i$'s dataset by creating $\mathcal{D}^\texttt{CP}_i = \mathcal{D}^\texttt{CP}_i \cup \{\hat{\bm{x}}, \hat{\bm{y}}^\texttt{CP}_i\}$. 
}
\caption{\small Constraint Primitive Bayesian Optimisation (CP-BO)}\label{alg:cpbo}
\end{algorithm}





\begin{figure}[t!]
    \centering
    
    \begin{subfigure}[b]{0.43\textwidth}
        \centering
        \includegraphics[width=1\textwidth]{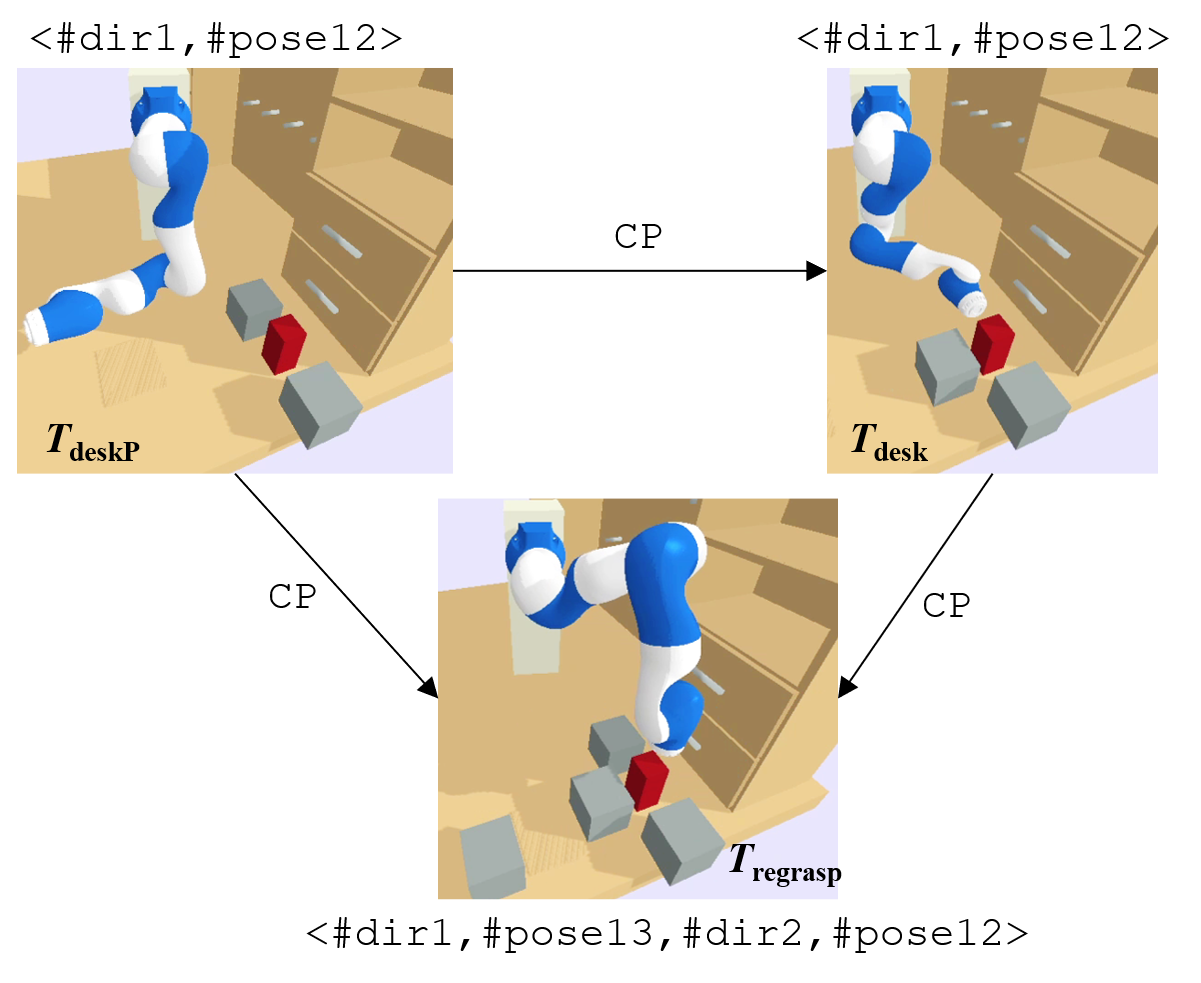}
            \vspace{-0.18cm}
        \caption{$T_\text{deskP}\rightarrow T_\text{desk}$,  $\{T_\text{deskP},T_\text{desk}\}\rightarrow T_\text{regrasp}$}
        \label{fig:transDesk}
    \end{subfigure}
    \hfill
    \begin{subfigure}[b]{0.51\textwidth}
        \centering
        \includegraphics[width=1\textwidth]{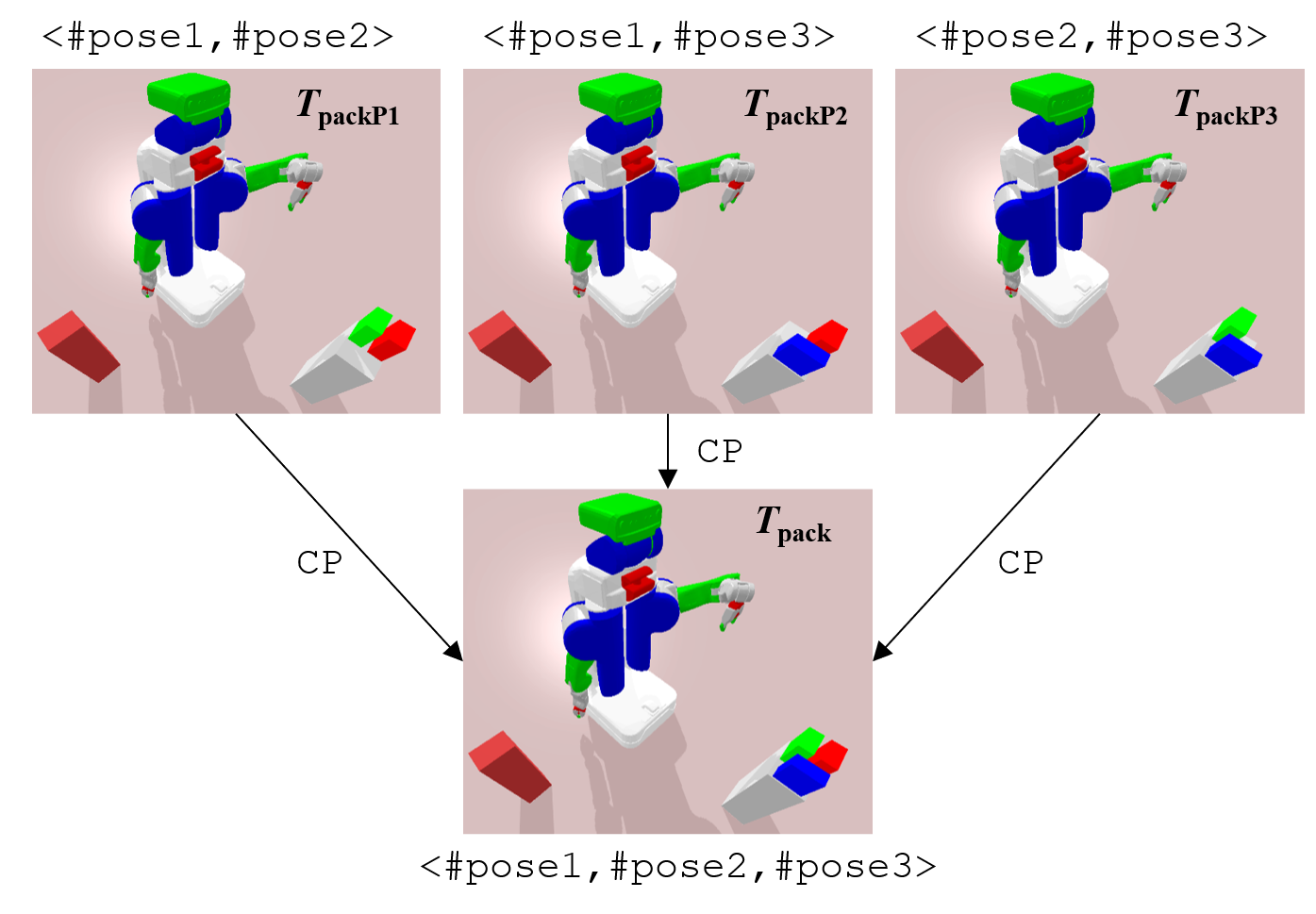}
            \vspace{-0.18cm}
        \caption{$\{T_\text{packP1},T_\text{packP2},T_\text{packP3}\}\rightarrow T_\text{pack}$}
        \label{fig:transPack}
    \end{subfigure}
    
    \caption{\small The constraint primitives transferred across different tasks, as indicated by the arrows. CP points relevant to the new tasks are automatically activated and utilized by CP-BO to assist binding search.}
    \label{fig:transTask}
    \vspace{-0.5cm}
\end{figure}


\subsection{Transferring Constraint Primitive Knowledge}\label{sec:methodsc}
In the binding search of another task, we encode each encountered failure as a type ID and use it to query in the global CP dictionary $\mathcal{D}$ (see Fig.\,\ref{fig:tranfer_cp} (right)). If it matches with existing IDs, certain groups of CP points are activated and connected to their casual decision tree nodes. Then, we can use these CP points to guide exploration at these decision tree nodes by CP-BO. We note this knowledge transfer mechanism by \textit{CP-BO Transfer}. \textit{CP-BO Transfer} is fully automated and its effectiveness depends on the amount of relevant knowledge already stored in $\mathcal{D}$.

\section{Empirical evaluation} 
\label{sec:evaluation}

\begin{figure}[t!]
        \centering
        \includegraphics[width=1.0\linewidth]{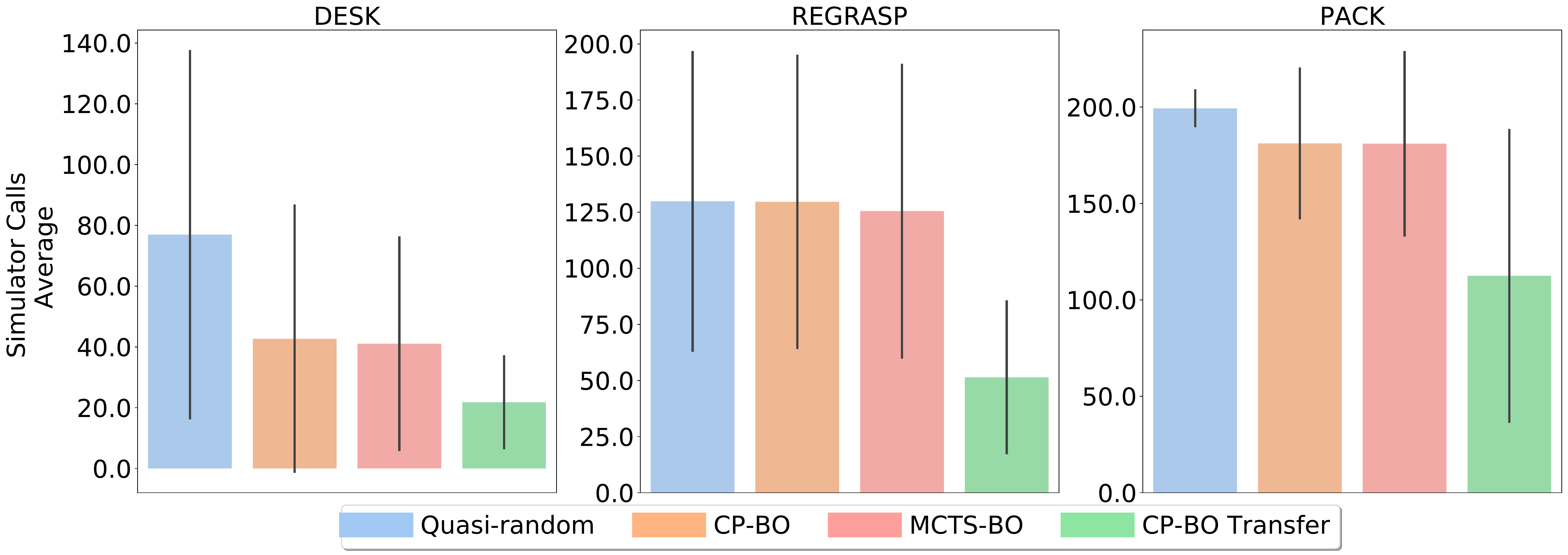}
    \caption{Evaluation results of the the baseline algorithm \texttt{Quasi-random} vs our three proposed algorithms \texttt{CP-BO}, \texttt{MCTS-BO} and \texttt{CP-BO Transfer} across three benchmark tasks. These box plots show the number of rollouts in the simulator required to achieve feasibility, with confidence intervals (+-sd). From these figures, we observe that both all our purposed \texttt{CP-BO}, \texttt{MCTS-BO}  versions of PW-UCT consistently outperform \texttt{Quasi-random} PW-UCT. Whilst \texttt{CP-BO Transfer} PW-UCT offers a significant further reduction in rollouts in the simulator.
    }
    \label{fig:results}
\end{figure}

The aim of this section is to evaluate the advantages of our framework for transfer learning in TAMP. We evaluate our proposed methods on three robot tasks: three desk task ($T_\text{desk}$ in Fig.\,\ref{fig:task_desk}) which has two binding decisions ($\langle \texttt{\#dir1,\#pose12} \rangle$ in Fig.\,\ref{fig:task_desk}), the re-grasping task ($T_\text{regrasp}$) which has four binding decisions, and the packing task ($T_\text{pack}$) which has three binding decisions~\footnote{See Fig.\,\ref{fig:transTask}. Please see Appendix\,\ref{app:b} for more details on tasks.}. We also make the following assumption:
\begin{assumption}\label{App:Assumption_1}
We assume that feasible skeletons are already given by symbolic planners, thus only require correct binding of variables.
\end{assumption}

\textbf{Binding Search Methods: }We evaluate four methods for guiding node expansion in PW-UCT, by assigning values to bindings. First, the baseline method, \textit{Quasi-random}, which uses a Voronoi sampler to bind variables and thus does not collect or utilise the geometric constraint primitive knowledge outlined in Sec.~\ref{sec:methods}. \textit{MCTS-BO} performs GP-UCB~\cite{srinivas2009gaussian} directly on the reward function~\ref{eq:reward} of PW-UCT, thus does not collect or utilise the geometric constraint primitive knowledge as its collected dataset of bindings and reward are dependent on the specific tasks success. Finally, we evaluate our proposed \textit{CP-BO} which constructs a dataset of geometric constraint primitive knowledge (Sec.~\ref{sec:methodsa}), and utilises it using BO~(Sec.~\ref{sec:methodsb}). Finally in \textit{CP-BO Transfer}, we allow \textit{CP-BO} predecessor tasks to collect CP points into their respective datasets $\mathcal{D}^\texttt{CP}_\texttt{typeID}$. To prevent \textit{CP-BO} from using previous knowledge, we reset $\mathcal{D}$ before each scenario. 

\textbf{Surrogate Model Design: }For modelling the continuous UCT reward or modelling the constraint primitives by a GP we use; a $\text{Mat\'{e}rn}(5/2)$ covariance function with automatic relevance detection and a constant mean function. Additionally, we apply SMOTE to tackle the issue of unbalanced classes in GP classification.


\textbf{Transfer Learning Setup: }For transfer learning, as shown in Fig.\,\ref{fig:transTask}, $T_\text{desk}$ has one predecessor task, $T_\text{regrasp}$ has two and $T_\text{pack}$ has three. Each predecessor task is evaluated for 30 rollouts. 

Each algorithm is evaluated on the benchmark tasks for 1000 seeds and we show the results in Fig.\,\ref{fig:results}. We observe that in most tasks that, even without transferring knowledge, reduces rollouts in the simulator: shown by \textit{Quasi-random} performing the worst compared with \textit{CP-BO} and \textit{MCTS-BO}. Finally, in \textit{CP-BO Transfer} we see significant reductions in rollouts required to find a feasible binding set. We observe a 71.69 \% relative reduction on expensive simulation calls vs the baseline \textit{Quasi-random} on $T_\text{desk}$, on $T_\text{regrasp}$ a 60.37\% relative reduction and a 43.60 \% relative reduction on $T_\text{pack}$. We verify via that \textit{CP-BO} acquired this gain is derived solely from \textit{CP-BO Transfer} representing, efficiently utilising and transferring CP knowledge across tasks.  


\section{Conclusion} 
\label{sec:conclusion}

Based on a novel representation of the geometric constraints in TAMP, we propose a CP-BO to guide the UCT exploration in binding search for solving challenging TAMP tasks. Our empirical evaluation shows its effectiveness in a variety of robot tasks. Especially when previous CP knowledge are available, the CP-BO solvers can generate suggestions with higher quality from limited data than the comparable task-dependant approach MCTS-BO.
With CP-BO, the planner is able to accumulate knowledge in a stream of tasks and to achieve a constant transfer learning of the geometric environment.

\appendix
\section{Additional Methodological background} \label{app:a}
\subsection{PW-UCT: MCTS for binding search}\label{app:a}
 In this study, we use a reward function defined as
$r = 0.1\left(d_\text{end}/d_\text{total}\right)+r_\text{success}$, where $d_\text{end}$ is the depth in UCT where $node$ is terminated. The first term of $r$ is a normalized depth that encourages the robot to avoid branches where bindings could fail earlier. For the second term, $r_\text{success}=1$ when all bindings are successfully found, otherwise $r_\text{success}=0$. By BACK-PROPAGATION$(node)$, the reward is assigned to $node$ and its parents as a standard MCTS procdeure.
In each rollout, ADD-VISIT$(node)$ increases the visit number of $node$ by 1 and meanwhile updates the environment state.  In a PW law, a sufficient visiting number means that the value of an existing node is estimated sufficiently well, then new nodes will be expanded to explore the unreached decision space. In addition, if the decision at $node$ is finite, PW-EXPANDABLE$(node)$ will prevent further node expansion after all choices are enumerated. For further details we refer the readers to \cite{ren2021extended}

\section{Experiment details} \label{app:b}

\subsection{The TAMP tasks for evaluation}

 \begin{figure*}[t!]
    \centering
        \begin{subfigure}[b]{0.5\linewidth}
        \centering
        \includegraphics[width=1\linewidth]{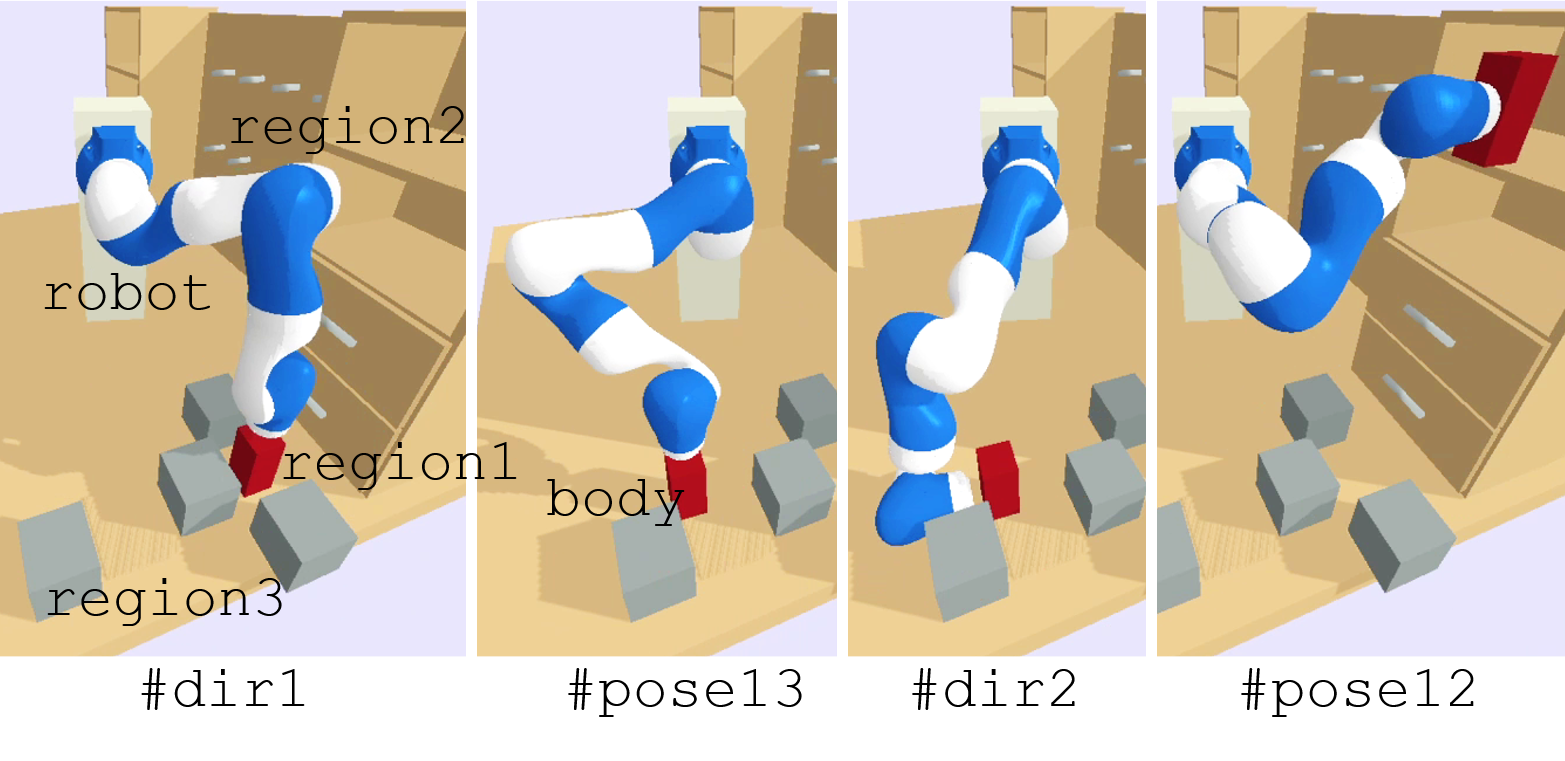}
        \caption{$T_\text{regrasp}$}
        \label{fig:task_regrasp}
    \end{subfigure}
    \hfill
    \begin{subfigure}[b]{.6\linewidth}
        \centering
        \includegraphics[width=1\linewidth]{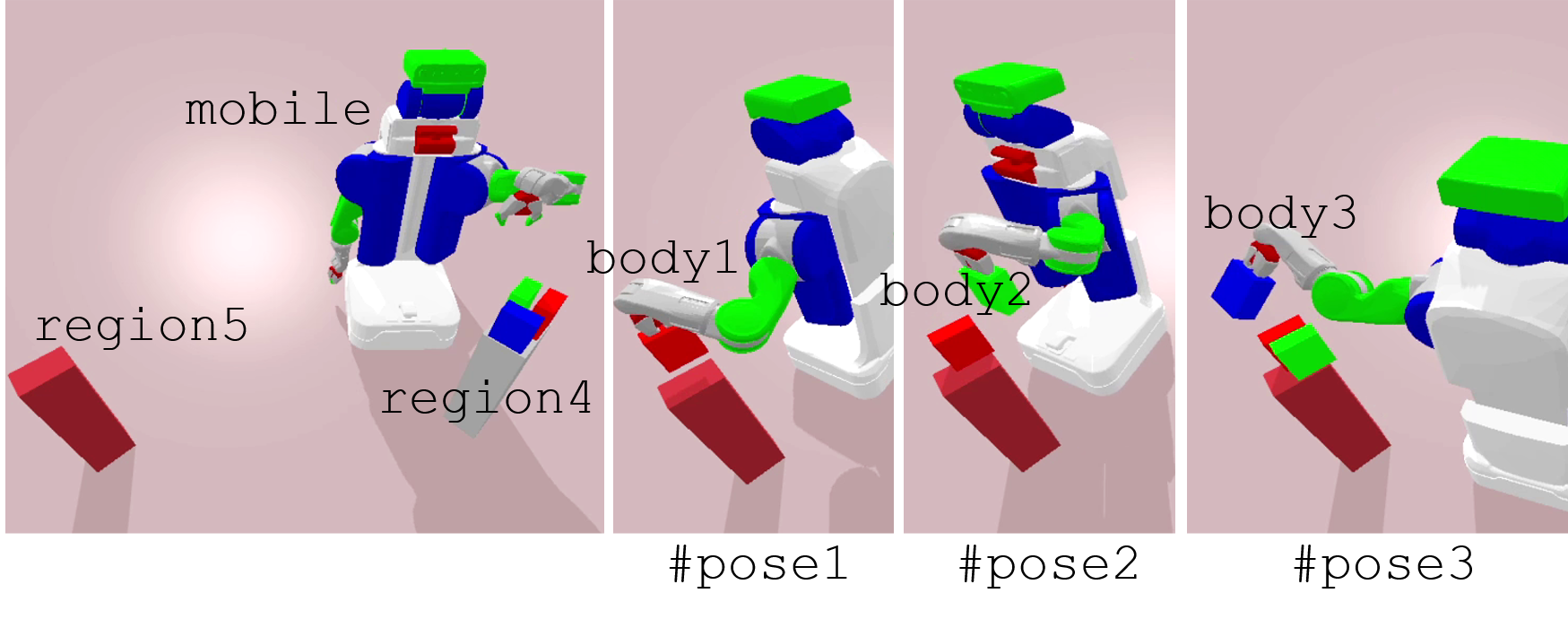}
        \caption{$T_\text{pack}$}
        \label{fig:task_pack}
    \end{subfigure}

    \caption{\small TAMP tasks for evaluation.}
    \label{fig:regrasp}
\end{figure*}

\textit{\underline{The regrasping task.}}
For $T_\text{regrasp}$ as shown in Fig.\,\ref{fig:task_regrasp}, we have the problem BINDING$(T_\text{regrasp},\langle \#dir1, \#pose13, \#dir2, \#pose12 \rangle)$ to solve. Compared to $T_\text{desk}$ in Fig.\,\ref{fig:task_desk}, $T_\text{regrasp}$ is more challenging in binding search since it has a longer horizon with two more decisions to make. To avoid collision with obstacles and meanwhile facilitate the following regrasping motion, \texttt{\#pose13} of \texttt{body} in \texttt{region3} should be carefully chosen. Then the planner should search for a new grasping direction \texttt{\#dir2} that enable a collision-free trajectory on the shelf later during the placement in \texttt{region2} while not colliding with obstacles on the table at the moment. 

\textit{\underline{The packing task.}}
In $T_\text{pack}$ from Fig.\,\ref{fig:task_pack}, the mobile robot must relocate \texttt{body1}, \texttt{body2}, \texttt{body3} from \texttt{region4} to \texttt{region5} without causing any collisions and drops of the movable bodies. The new locations of all the bodies in \texttt{region5}: \texttt{\#pose1}, \texttt{\#pose2}, \texttt{\#pose3}, should be strategically decided to ensure that the first-arriving bodies don't occupy too much space. To make the task even more difficult, we set \texttt{region5} 25\% smaller than \texttt{region4}. This binding search problem is noted as BINDING$(T_\text{pack},\langle \#pose1,\#pose2,\#pose3 \rangle)$. 

\subsection{CP-BO with transferred knowledge} 

As shown in Fig.\,\ref{fig:transTask}, the predecessor task of $T_\text{desk}$ is $T_\text{deskP}$ while $T_\text{regrasp}$ have two predecessor tasks, $T_\text{deskP}$ and $T_\text{regrasp}$. Similarly, we run three predecessor tasks including $T_\text{packP1}$, $T_\text{packP2}$, $T_\text{packP3}$ are ahead of $T_\text{pack}$. Each predecessor task is evaluated for 30 rollouts. We do not expect that the surrogate models of CP-BO is fully converged within such data-collection sessions, but give the Trans-CP-BO algorithm a warm start.

\bibliography{LaTeX/aaai22}
\section{Acknowledgments}
....

\end{document}